\newcommand{\change}[1]{\textcolor{black}{#1}}
\newcommand{\changefinal}[1]{\textcolor{black}{#1}}
\documentclass{article}
\usepackage{lineno}
  \usepackage[pdftex]{graphicx}
  \usepackage{caption}
\usepackage{subcaption}
\usepackage{xcolor}
  \DeclareGraphicsExtensions{.pdf,.jpeg,.png}
\usepackage{amsmath, amsthm, amssymb}
\usepackage{array}
\usepackage{url}

\begin{document}
\newcommand{\TODO}[1]{{\textcolor{red}{#1}}}
\newcommand{\CHANGEFS}[1]{{\textcolor{orange}{#1}}}

\newcommand{\x}{{\bf x}}
\newcommand{\w}{{\bf w}}
\newcommand{\uu}{{u}}
\newcommand{\uul}{{{\bf u}_\ell}}
\newcommand{\ff}{{\bf f}}
\newcommand{\ffl}{{\bf f}_\ell}
\newcommand{\uull}{{\bf u}_{\ell+1}}
\newcommand{\UUl}{{\hat{\bf u}}_{\ell}}
\newcommand{\UUll}{{\hat{\bf u}}_{\ell+1}}
\newcommand{\act}{{\varphi}}
\newcommand{\UU}{{\hat{u}}}
\newcommand{\dims}{{N_D}}
\newcommand{\nodes}{\changefinal{N}}
\newcommand{\dimout}{{N_O}}
\newcommand{\dimint}{{N_I +1}}
\newcommand{\dimintred}{{N_I}}
\renewcommand{\d}[1]{\ensuremath{\operatorname{d}\!{#1}}}
\newtheorem{theorem}{Theorem}
\newcommand{\g}{{g}}
\newcommand{\G}{{\hat{g}}} 
\newcommand{\FF}{{\hat{f}}} 
\newcommand{\FFl}{{\hat{{\bf f}}_\ell}} 
\newcommand{\errwj}{{e_{{\bf u}_j}}}
\newcommand{\err}{{e}}
\newcommand{\erru}{{{\bf e}_u}}
\newcommand{\lipact}{{l_{\varphi}}}
\newcommand{\lipg}{{l_{g}}}
\newcommand{\lipl}{{l_\ell}}
\newcommand{\lipll}{{l_{\ell+1}}}

%
\title{\change{Learning with Known Operators reduces Maximum Training Error Bounds}}


\author{Andreas~K.~Maier,
        Christopher~Syben, Bernhard~Stimpel, \\
        Tobias W{\"u}rfl,
        Mathis Hoffmann,
        Frank Schebesch,
        Weilin~Fu,
        \\Leonid~Mill\thanks{A.K. Maier, C. Syben, T  W{\"u}rfl, M. Hoffmann, F. Schebesch, W. Fu, B. Stimpel, and L. Mill are with the Department of Computer Science, Friedrich-Alexander University Erlangen-N{\"u}rnberg, Germany. E-mail: andreas.maier@fau.de.},
 ~Lasse Kling\thanks{L. Kling is with the Helmholtz Zentrum Berlin f\"ur Materialien und Energie, Germany.}, and~Silke~Christiansen\thanks{S.Christiansen is with the Physics Department, Free University Berlin and the Helmholtz Zentrum Berlin f\"ur Materialien und Energie, Germany.}
}


%
%




\begin{abstract}

We describe an approach for incorporating prior knowledge \change{into machine learning algorithms}. We aim at applications in physics and signal processing in which we know that certain operations must be embedded into the algorithm. Any operation that allows computation of a gradient or sub-gradient towards its inputs is suited for our framework. 
\changefinal{We derive a maximal error bound for deep nets that demonstrates that inclusion of prior knowledge results in its reduction.}
Furthermore, we \changefinal{also show}  experimentally that \changefinal{known operators reduce} the number of free parameters. We apply this approach to various tasks ranging from CT image reconstruction over vessel segmentation to the derivation of previously unknown imaging algorithms. As such the concept is widely applicable for many researchers in physics, imaging, and signal processing. We assume that our analysis will support further investigation of \change{known operators} in other fields of physics, imaging, and signal processing.
\end{abstract}


\maketitle


%

\section{Introduction}
%
%
%
%
Pattern analysis and machine intelligence have been focussed predominantly on tasks that mimic perceptual problems. These are typically modelled as classification or regression tasks in which the actual reference stems from a human observer that defines the {\it ground-truth}. As we have only limited understanding on how these man-made classes emerge from the human mind, there is only limited knowledge available. As such, pattern recognition has relied on expert knowledge to design features that are suited towards a particular recognition task \cite{niemann2013pattern}. In order to alleviate the task of feature-design, researchers started also learning feature descriptors as a part of the training procedure \cite{lecun1995convolutional}. Implementation of such on efficient hardware gave rise to first models that could outperform classical feature extraction methods significantly \cite{krizhevsky2012imagenet} and was one of the milestone works in the emerging field of {\it deep learning}. 

With the rise of {deep learning} \cite{lecun2015deep}, researchers became aware that these methods of general function learning are applicable to a much wider range than mere perceptual tasks. Today, machine learning is applied in a much wider range of applications. Examples range from image super resolution \cite{dong2014learning}, image denoising and inpainting \cite{xie2012image}, or even computed tomography \cite{wang2018image}. In these fields, the methods from deep learning are often directly applied and often show performances that are either en par or even significantly better than results found with state-of-the-art methods. Yet, there are also reports that present surprising results in which parts of the image are hallucinated \cite{Cohen2018distribution, huang2018considerations}. In particular \cite{huang2018considerations} demonstrates that mismatches in training and test data leads to dramatic changes in the produced result. Hence, {\it blind} deep learning methods have to be performed with care in order to be successful. 

In this article, we explore the use of known operations within machine learning algorithms. First, we analyze the problem from a theoretical perspective and study the effect of prior knowledge in terms of maximal error bounds. This is followed by three applications in which we use prior operators to study to their effect on the respective regression or classification problem. Lastly, we discuss our observations in relation to other works in literature and give an outlook on future work. Note that some of the work presented here is based on prior conference publications \cite{deeplearningct, ArXivWeilin, maier2018precision, syben2018deriving}.

\section{Known Operator Learning}

\begin{figure*}[tbp]
\begin{center}
\includegraphics[width=\linewidth]{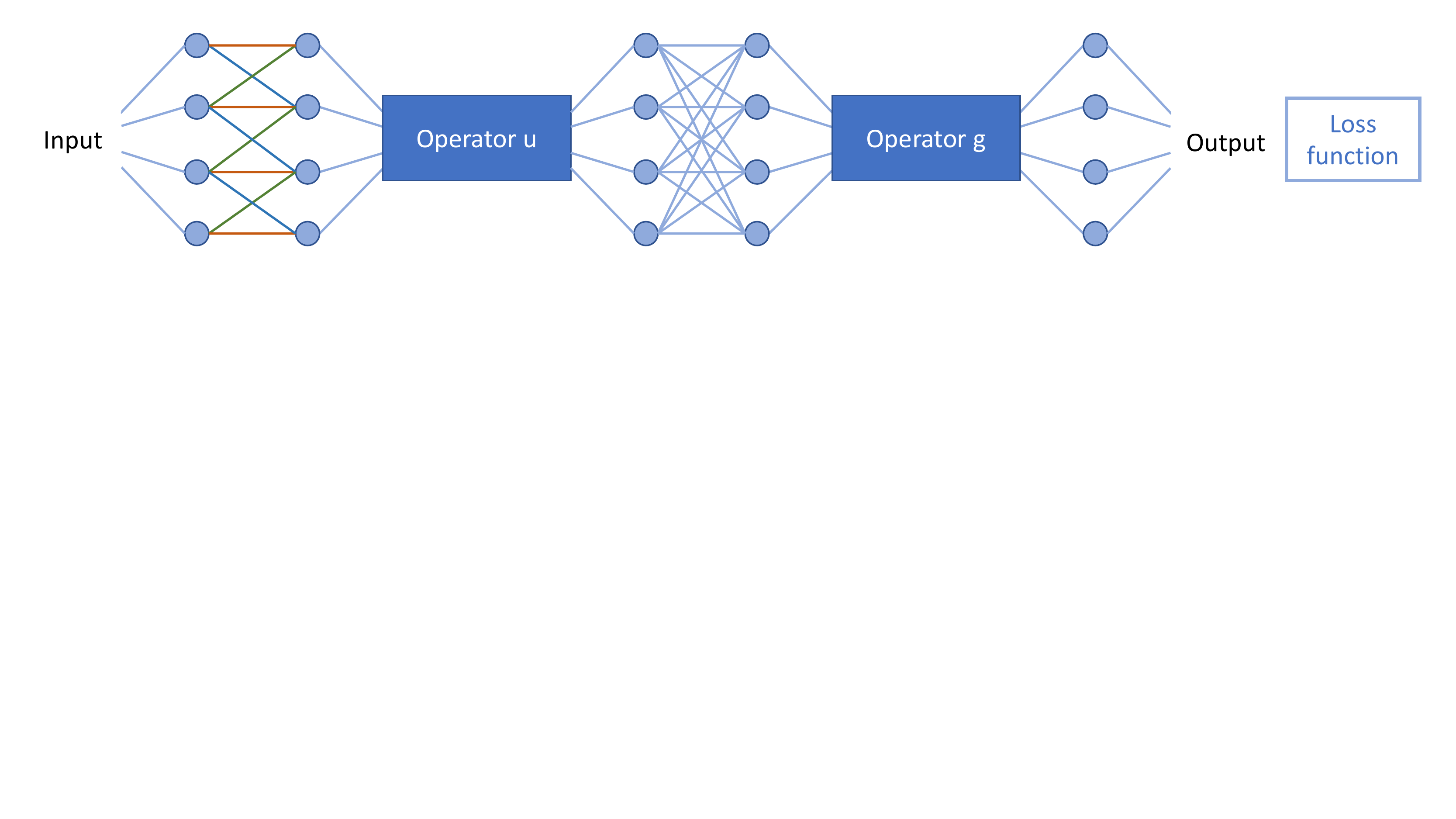}\vspace{-4.5cm}
\caption{Schematic of the idea of known operator learning: One or more known operators \change{(here Operator u and Operator g)} are embedded into a network. Doing so, allows dramatic reduction of the number of parameters that have to be estimated during the learning process. The minimal requirement for the operator is that it must allow the computation of a sub-gradient for use in combination with the back-propagation algorithm. This requirement is met by a large class of operations.}
\label{fig:precision}
\end{center}
\end{figure*}

The general idea of known operator learning is to embed entire operations into a learning problem. Figure~\ref{fig:precision} presents the idea graphically. We generally refer to the ${\dims}$-dimensional input of our trained algorithm as $\x' \in \mathbb{R}^{\dims}$. In order to increase readability, we use an extended version $\x \in \mathbb{R}^{\dims+1}$ such that inner products with some weight vector $\bf w'$ plus bias $w_0$ can be conveniently written, i.\,e. ${\bf w'}^\top {\bf x'} + w_0 = {\w}^\top{\x}$. Before looking into the properties of this approach and in particular maximal error bounds, we shortly summarize the Universal Approximation Theorem as it is closely related to our analysis. Note that the supplementary material to this article contains all proofs for the presented theorems in this section.

\subsection{Universal Approximation}

\begin{theorem}[Universal Approximation Theorem]
Let $\act(t):\mathbb{R} \rightarrow \mathbb{R}$ be a non-constant, bounded, and continuous function and $\uu(\x)$ be a continuous function on a compact set $\mathcal{D} \subset \mathbb{R}^{\dims+1}$. Then, there exist an integer $\nodes$, weights $\w \in \mathbb{R}^{\dims+1}$, and $u_i \in \mathbb{R}$ that form an approximation $\UU(\x)$
\begin{equation}
\UU(\x) = \sum_{i=0}^{\nodes-1} u_i \act({\bf w}_i^\top {\bf {x}})
\end{equation} 
 such that the inequality 
\begin{equation}
|\UU(\x) - u(\x)| \le \epsilon_u
\end{equation}
holds for all $\x \in \mathcal{D}$ and $\epsilon_u > 0$.
\label{theo:universal}
\end{theorem}

Theorem~\ref{theo:universal} states that for any continuous function $\uu(\x)$ an approximation $\UU(\x)$ can be found such that the difference between true function and approximation is bounded by $\epsilon_u$. With increasing number of nodes $\nodes$, $\epsilon_u$ will decrease. In literature, this result is often referred to as Universal Approximation Theorem \cite{cybenko1989approximation, hornik1991approximation} and forms the fundamental result that neural networks with just a single hidden layer are general function approximators. Yet, this type of approximation may result in a very high requirement for the choice of $\nodes$ which is the reason why stacked layers of different type are known to be more successful \cite{maier2019gentle}.

\renewcommand{\uu}{{\bf u}}
\renewcommand{\UU}{{\bf \hat{u}}}

We can extend Theorem~\ref{theo:universal} to vector-valued functions $\uu(\x): \mathcal{D} \rightarrow \mathbb{R}^{\dimout}$ on $\mathcal{D}$ by postulating Theorem~\ref{theo:universal}  for each of their components $k$
\begin{equation}
|U_k(\x) - u_k(\x)| \le \epsilon_{u_k}.
\label{eq:generalisation}
\end{equation}
Hence, universal approximation generally also applies to $\dimout$-dimensional functions.

\subsection{Known Operator Error Bounds}

Knowing the limits of general function approximation, we are now interested in finding limits for mixing known and approximated operators. As previously mentioned, deep networks are never constructed out of a single layer, but rather take the form of the configuration shown in Figure~\ref{fig:precision}. Hence, we need to consider layered networks to analyze the maximal error bounds. Instead of investigating entire networks, we choose to simplify our theoretical analysis to the special case 
$$f(\x)= \g(\uu(\x)): \mathcal{D} \rightarrow \mathbb{R}$$
with 
{$\g(\x):S \rightarrow\mathbb{R}$, $\uu(\x):\mathcal{D} \rightarrow \mathcal{S}$}, and compact sets $\mathcal{S} \subset \mathbb{R}^\dimint$ and $\mathcal{D} \subset \mathbb{R}^{\dims+1}$. Note that this simplification does not limit the generality of our analysis, as we can map any knowledge on the structure of the network architecture either onto the output function $\g(\x)$, the intermediate function $\uu(\x)$, or directly as a transform of the inputs $\x$. Generalisation to $\dimout$-dimensional functions is also possible following the idea shown in Eq.~\ref{eq:generalisation}.

Previous definition of $f(\x)$ allows us to investigate different forms of approximation. In particular, we are able to introduce approximations $\UU(\x)$ and $\G(\x)$ following Theorem~\ref{theo:universal}:

\begin{eqnarray}
\FF_g(\x)=&\G(\uu(\x))&= f(\x)-\err_g \label{eq:eg}\\
\FF_u(\x)=&\g(\UU(\x)) &= f(\x)-\err_u \label{eq:eu}\\
\FF(\x)=&\G(\UU(\x)) &= f(\x)-\err_f \label{eq:ef}
\end{eqnarray}
Here $|\err_u| \le \epsilon_u$, $|\err_\g| \le \epsilon_\g$, and $|\err_f| \le \epsilon_f$  denote the errors that are introduced by respective approximation of $\uu$, $\g$, and $f$.

Next, we are interested in finding bounds on $|e_f|$ using above approximations. For the case of known $\uu(\x)$, we can substitute $\x^\ast:=\uu(\x)$, as $\uu(\x)$ is a fixed function. In this case Theorem~\ref{theo:universal} directly applies and a bound on $|e_f|$ is found as  $|e_f| \leq \epsilon_g$ with $|e_g| \leq \epsilon_g$. If we would know $\g(\x)$ in addition, $e_g$ would be 0 and the bound would shrink to the case of equality.

The case described in Eq.~\ref{eq:eu} is slightly more complicated, but we are also able to find general bounds as shown in Theorem~\ref{theo:known2}.

\begin{theorem}[Known Output Operator Theorem]
Let $\act(\x):\mathbb{R} \rightarrow \mathbb{R}$ be a non-constant, bounded, and continuous function and $f(\x)=g(\uu(\x)):\mathcal{D} \rightarrow \mathbb{R}$ be a continuous function on $\mathcal{D} \subset \mathbb{R}^{\dims+1}$. Further let $g(\x): \mathcal{S} \rightarrow \mathbb{R}$ be Lipschitz-continuous function with Lipschitz constant $\lipg = \textnormal{sup}\{||\nabla g(\x) ||_p\}$ with $p\in \{1,2\}$ on $\mathcal{S} \subset \mathbb{R}^\dimint$ and 
\begin{eqnarray}
{\hat{u}}_k(\x) &=&  \sum_{i=0}^{N_{\hat{u}_k}-1} u_i \act({\bf w}_{i,k}^\top {\bf {x}})
\end{eqnarray}
be a general function approximator of $\uu(\x)$ with integer $N_{\hat{u}_k}$, weight $\w_{i,k} \in \mathbb{R}^{\dims+1}$, and $u_i \in \mathbb{R}$. Then, $e_f = f(\x) - \FF(\x)$ with $\FF(\x) = \G(\UU(\x))=\g(\UU(\x))$ ---~as $\g$ is known ---  is generally bounded for all $\x \in \mathcal{D}$ by 
\begin{equation}
|e_f| \le \lipg \cdot {||{\erru}||}_p \label{eq:boundg}
\end{equation}
with $e_u = e_f$ and component-wise approximation errors 
${\erru} = [{e}_{u_{0}}, \ldots e_{u_{\dimintred}}]^\top$.
\label{theo:known2}
\end{theorem}

The bound for $|e_f|$ is found using a Lipschitz constant $\lipg$ on $\g(\x)$ which implies that the theorem will only hold, if Lipschitz-bounded functions are used for $\g(\x)$. Analysis of Eq.~\ref{eq:boundg} reveals that knowing $\uu(\x)$ in this case, would imply $\erru = {\bf 0}$ which also yields equality on both sides.

\change{We further explore this idea in Theorem~\ref{theo:known1}. It describes a bound for the case that both $\g(\x)$ and $\uu(\x)$ are approximated. }

\begin{theorem}[Unknown Operator Theorem]
Let $\act(\x):\mathbb{R} \rightarrow \mathbb{R}$ be a non-constant, bounded, and continuous function with Lipschitz-bound $\lipact$ and $f(\x)=\g(\uu(\x)):\mathcal{D} \rightarrow \mathbb{R}$ be a continuous function  on $\mathcal{D} \subset \mathbb{R}^{\dims+1}$. Further let 
\begin{eqnarray}
\G(\x) &=&  \sum_{j=0}^{N_\G-1} g_j \act({\bf w}_j^\top {\bf {x}}) \quad\textnormal{and}\\
{\hat{u}}_k(\x) &=&  \sum_{i=0}^{N_{\hat{u}_k}-1} u_i \act({\bf w}_{i,k}^\top {\bf {x}})
\end{eqnarray}
be general function approximators of $g(\x): \mathcal{S} \rightarrow \mathbb{R}$ and $\uu(\x): \mathcal{D} \rightarrow \mathcal{S}$ with integers $N_\G$ and $N_{\hat{u}_k}$, weights $\w_j \in \mathbb{R}^\dimint$, $\w_{i,k} \in \mathbb{R}^{\dims+1}$, $g_j, u_i \in \mathbb{R}$, and compact sets $\mathcal{S} \subset \mathbb{R}^\dimint$ and $\mathcal{D} \subset \mathbb{R}^{\dims+1}$. Then, $e_f = f(\x) - \FF(\x)$ with $\FF(\x) = \G(\UU(\x))$ is generally bounded for all $\x \in \mathcal{D}$ by 
\begin{equation}
|e_f| \le \sum_{j=0}^{N_\G-1} |g_j| \cdot \lipact \cdot |\errwj| + \epsilon_g.
\end{equation}
where $\epsilon_g \geq |\err_g|$, $\errwj=\w_j^\top {\erru}$, and $\erru= [{e}_{u_{0}}, \ldots e_{u_{\dimintred}}]^\top$ is the vector of errors introduced by the components of $\UU(\x)$.
\label{theo:known1}
\end{theorem}

The bound is comprised of two terms in an additive relation:
\begin{equation}
\underbrace{\sum_{j=0}^{N_\G-1} |g_j| \cdot \lipact \cdot |\errwj|}_{\text{only dependent on } \UU(\x)} + \underbrace{\epsilon_g}_{\text{only dependent on } \G(\x)}
\end{equation}
where the first term vanishes, if $\uu(\x)$ is known as $|\errwj| = 0~ \forall j$ and the second term vanishes for known $\g(\x)$ as $e_g = 0$. 
Hence for all of the considered cases, knowing $\g(\x)$ or $\uu(\x)$ is beneficial and allows to shrink the maximal training error bounds.

\renewcommand{\FF}{{\bf \hat{f}}} 

\change{
Given the previous observations, we can now also explore deeper networks that try to mimic the structure of the original function. This gives rise to Theorem~\ref{theo:known1deep}.}

\change{
\begin{theorem}[Unknown Operators in Deep Networks]
Let $\uul(\x_\ell):\mathcal{D}_{\ell} \rightarrow \mathcal{D}_{\ell-1}$ be a continuous function with Lipschitz-bound $l_\uul$ on compact set $\mathcal{D}_\ell \subset \mathbb{R}^{N_\ell}$ with integer $\ell > 0$. Further let $\ffl(\x_\ell): \mathcal{D}_\ell \rightarrow \mathcal{D}$ be a function composed of $\ell$ layers~/ function blocks defined as recursion $\ffl(\x_\ell) = \ff_{\ell-1}(\uul(\x_\ell))$ with $\ff_{\ell=0}(\x) = \x$ on compact set $\mathcal{D} \subset \mathbb{R}^{\dims+1}$ bound by Lipschitz constant $l_{\ffl}$ with $l_{\ff_{\ell=0}} = 1$. Recursive function $\FFl(\x_\ell) = \FF_{\ell-1}(\UU_{\ell}(\x_\ell))$ with $\FF_{\ell=0}(\x) = \x$ is then an approximation of $\ffl(\x_\ell)$. 
Then, ${\bf e}_{f, \ell} = \ffl(\x_\ell) - \FFl(\x_\ell)$ is generally bounded for all $\x_\ell \in \mathcal{D}_\ell$ and for all $\ell > 0$ in each component $k$ by 
\begin{equation}
|e_{f,\ell,k}| \le \sum_{\ell_i=1}^{\ell} ||\erru_{, \ell_i}||_p \cdot  l_{{\ff}_{\ell_i-1}}
\end{equation}
where 
$\erru_{,\ell}= [{e}_{u,\ell,{0}}, \ldots e_{u,\ell,{\dimintred}}]^\top$ is the vector of errors introduced by  $\UU_{\ell}(\x_\ell)$.
\label{theo:known1deep}
\end{theorem}
}

\change{
If we investigate Theorem~\ref{theo:known1deep} closely, we identify similar properties to Theorem~\ref{theo:known1}. The errors of each layer~/ function block $\uul(\x_\ell)$ are additive. If a layer is known, the respective error vector $\erru_{,\ell} := {\bf 0}$ vanishes and the respective part of the bound cancels out.
Furthermore, later layers have a multiplier effect on the error as their Lipschitz constants amplify $||\erru_{,\ell}||_p$. Note that the relation $l_{\ffl} \le \prod_{\ell_i=1}^{\ell} l_{\uul_i}$ is shown in the supplementary material. A large advantage of Theorem~\ref{theo:known1deep} over Theorem~\ref{theo:known1} is that the Lipschitz constants $l_{\ffl}$ that appear in the error term $||\erru_{, \ell_i}||_p \cdot  l_{{\ff}_{\ell_i-1}}$ are the ones of the true function $\ffl(\x_\ell)$. Therefore, the amplification effects are only dependent of the structure of the true function and independent of the actual choice of the universal function approximator. The approximator only influences the actual error $\erru_{, \ell_i}$.
}

Above observations pave the way to incorporating prior operators into different architectures. In the following, we will highlight several applications in which we explore blending deep learning with prior operators.

\section{Application Examples}

We believe that known operators have a wide range of applications in physics and signal processing. Here, we highlight three approaches to use such operators. All three applications are from the domain of medical imaging, yet the method is applicable to many more disciplines to be discovered in the future. The results presented here are based on conference contributions \cite{deeplearningct, ArXivWeilin, syben2018deriving}. Note that the supplementary material contains descriptions of experiments, data, and additional figures that were omitted here for brevity.

\subsection{Deep Learning Computed Tomography}
\label{sec:dlct}

\newcommand{\proj}{{\bf A}}
\newcommand{\y}{{\bf y}}
\newcommand{\fft}{{\bf F}}
\newcommand{\ifft}{{\bf F}^H}
\newcommand{\conv}{{\bf K}}
\newcommand{\convfft}{{\bf C}}
\newcommand{\weights}{{\bf W}}

In computed tomography, we are interested in computing a reconstruction $\y$ from a set of projection images $\x$. Both are related by the X-ray transfrom $\proj$:
$$\proj \y = \x$$
Solving for $\y$ requires inversion of above formula. The Moore-Penrose inverse of $\proj$ yields the following solution:
$$\y = \proj^\top(\proj\proj^\top)^{-1} \x$$
This type of inversion gives rise to the class of filtered back-projection methods, as it can be shown that $(\proj\proj^\top)^{-1}$ takes the form of a circulant matrix $\conv$, i.\,e. $\conv=(\proj\proj^\top)^{-1} = \ifft \convfft \fft$, where $\fft$ denotes the Fourier transform, $\ifft$ its inverse, and $\convfft$ a diagonal matrix that corresponds to the Fourier transform of $\conv$. As $\conv$ typically is associated with a large receptive field, it is typically implemented in Fourier space. In order to be applicable for other geometries, such as fan-beam reconstruction additional Parker and cosine weights have to be incorporated that can elegantly be summarised in an additional diagonal matrix $\weights$ to yield
\begin{equation}
\y = \text{ReLu}(\proj^\top \conv \weights \x) \label{eq:fbp}
\end{equation}
where $\text{ReLu}(\cdot)$ suppresses negative values as the final reconstruction algorithm.

\begin{figure}[tbp]
\begin{center}
\includegraphics[width=1\linewidth]{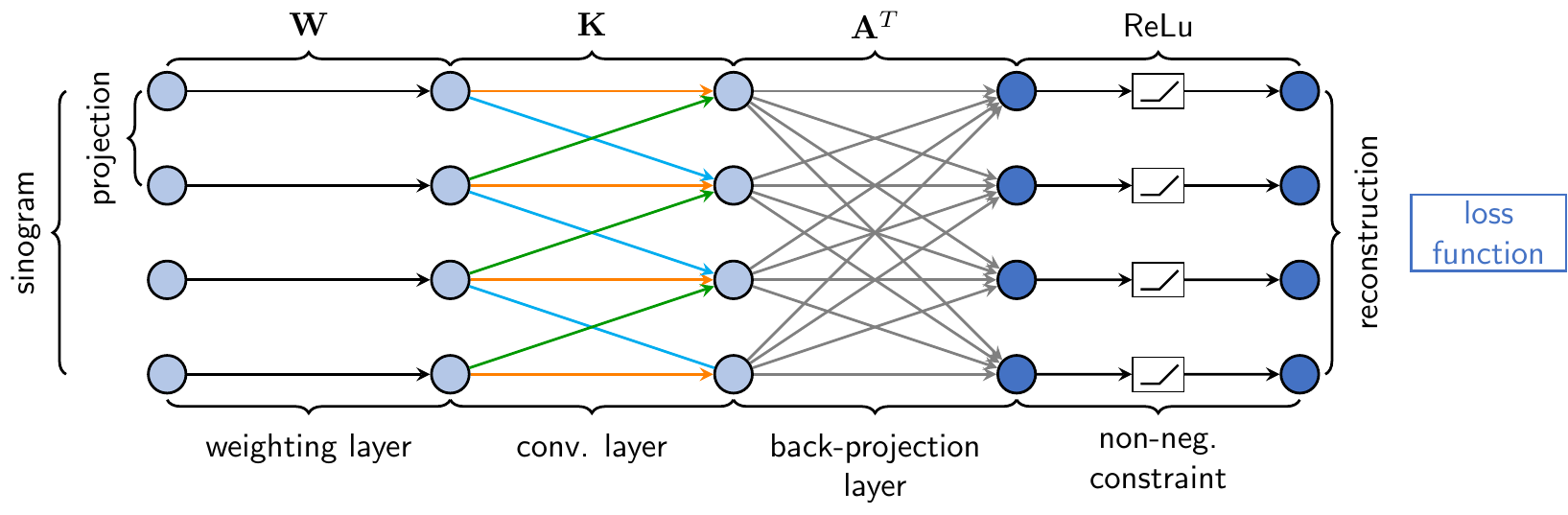}
\caption{\changefinal{Deep Learning Computed Tomography:} Reconstruction network for $\y = \text{ReLu}(\proj^\top \conv \weights \x)$ \change{from projections $\x$ to image $\y$}. As $\weights$ is a diagonal matrix, it is merely a point-wise multiplication followed by convolution $\conv$ and back-projection $\proj^\top$.}
\label{fig:fannet}
\end{center}
\end{figure}

Following the paradigm of known operator learning, Eq.~\ref{eq:fbp} can also interpreted as a neural network structure as it only contains diagonal, circulant, and fully connected matrix operators displayed in Figure~\ref{fig:fannet}. A practical limitation of $\proj$ is that it typically is a very large and sparse matrix. In practice, it is therefore never instantiated, but only evaluated on the fly using fast ray-tracing methods. For 3-D problems, the full matrix size is way beyond the memory restrictions of today's compute systems. Furthermore, none of the parameters need to be trained as all of them are known for complete data acquisitions. 

Incomplete data cannot be reconstructed with this type of algorithm and would lead to strong artifacts. 
We can still tackle limited data problems if we apply additional training of our network. As $\proj^\top$ is large, we treat it as fixed during the training and only allow modification of $\weights$ and $\conv$. Results and experimental details are demonstrated in the supplementary material. Training of both matrices clearly improves the image reconstruction result. In particular, the trained algorithm learns to compensate for the loss of mass in areas of the reconstruction in which rays are missing.

\begin{figure*}[tbp]
\begin{center}
\begin{subfigure}{0.33\linewidth}
  \centering
  \includegraphics[width=0.95\linewidth]{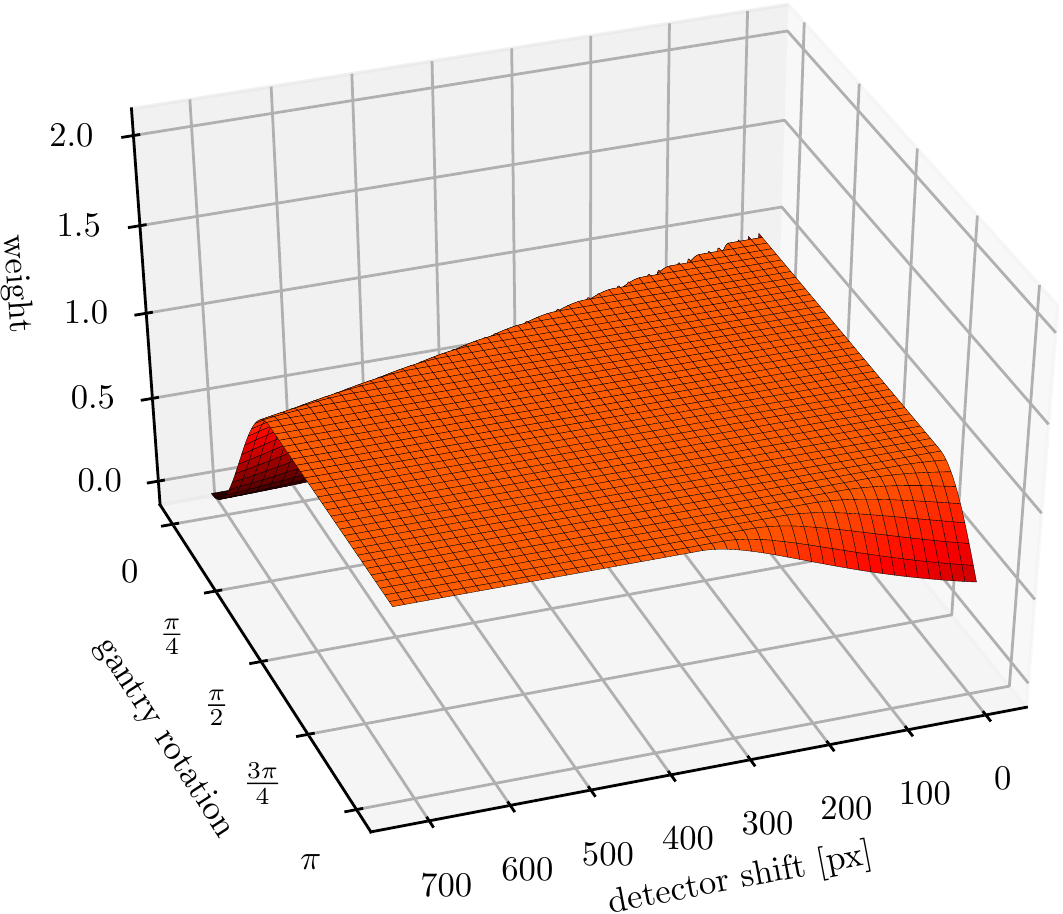}
  \caption{Parker Weights}
\end{subfigure}%
\begin{subfigure}{.33\linewidth}
  \centering
  \includegraphics[width=0.95\linewidth]{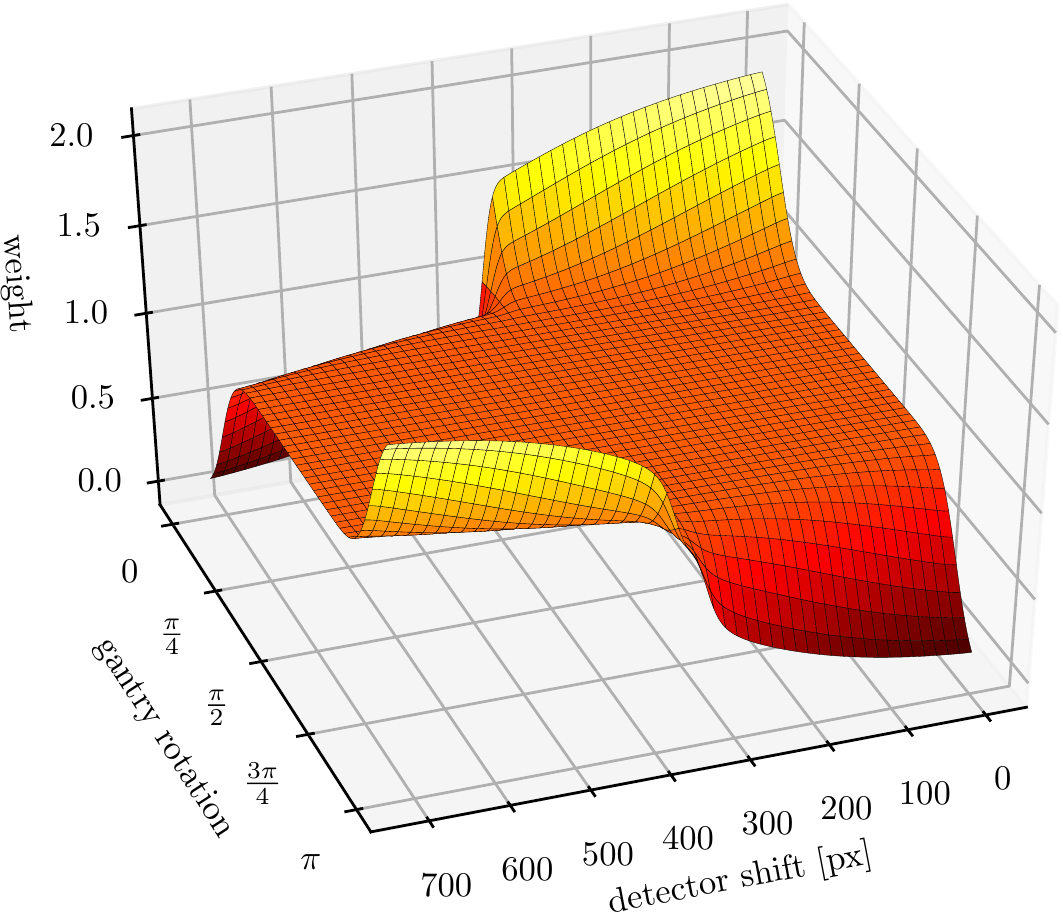}
  \caption{Sch{\"a}fer Weights}
\end{subfigure}%
\begin{subfigure}{.33\linewidth}
  \centering
  \includegraphics[width=0.95\linewidth]{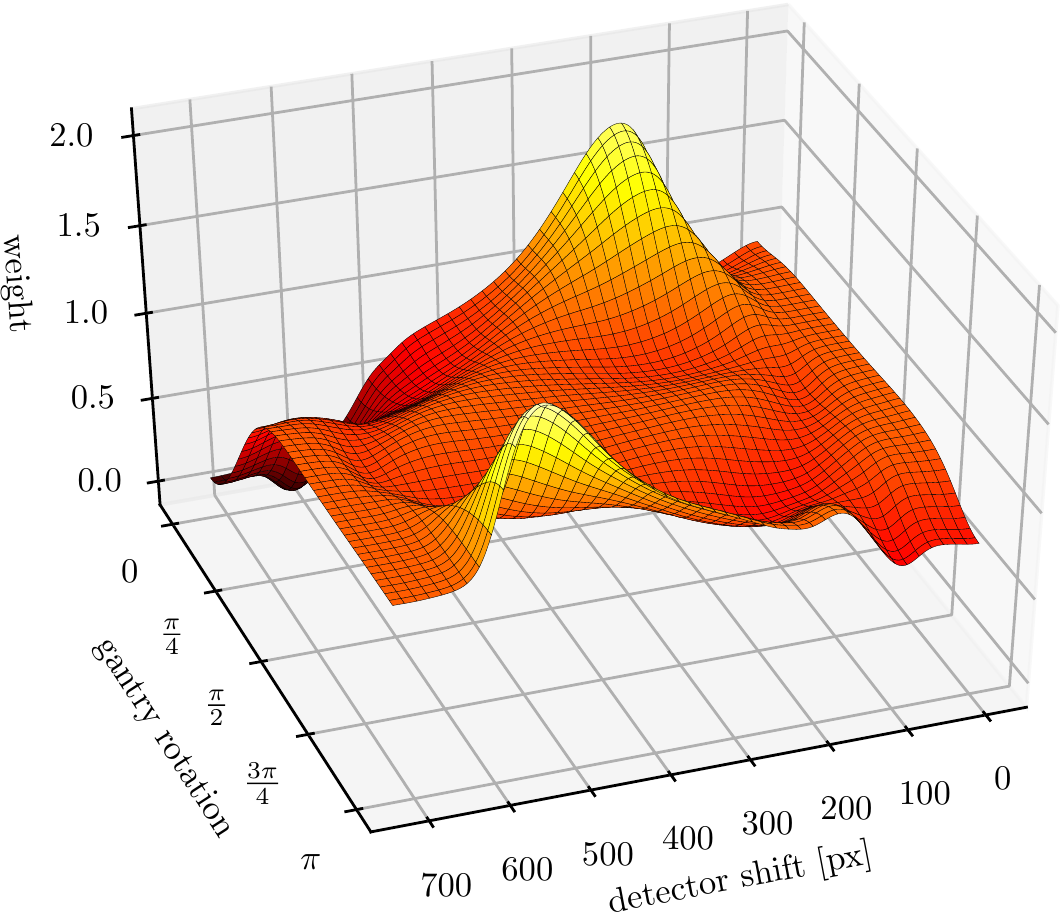}
  \caption{Trained Weights}
\end{subfigure}%
\caption{\changefinal{Improved interpretability in deep networks:} The trained reconstruction algorithm can be mapped back into its original interpretation. Hence, we can compare them to reconstruction weights after (a) Parker \cite{parker1982optimal} and (b) Sch{\"a}fer \cite{schafer2017modified}. (c) expresses significant similarity to (b) which is also able to compensate for the loss of mass. While (b) was only arrived at heuristically (c) can be shown to be data optimal here.}
\label{fig:redweights}
\end{center}
\end{figure*}

As the trained algorithm is mathematically equivalent to the original filtered back-projection method, we are able to map the trained weights back onto their original interpretation which allows comparison to state-of-the-art weights. In Figure~\ref{fig:redweights}, we can see that the trained weights show similarity with the approach published by Sch{\"a}fer et al. \cite{schafer2017modified}. In contrast to Sch{\"a}fer et al. who arrived at their weights following intuition, our approach is optimal with respect to our training data. \change{In our present model, we have to re-train the algorithm for every new geometry. This could be avoided by modelling the weights using a continuous function which is sampled by the reconstruction network.}


\subsection{Learning from Heuristic Algorithms}
\label{sec:vessel}

Incorporating known operators generally allows blending of deep learning methods with traditional image processing approaches. In particular, we are able to choose heuristic methods that are well understood and augment them with neural network layers. 

One example for such a heuristic method is Frangi's vesselness \cite{vesselness-frangi}. The vesselness values for dark tubes are calculated using the following formula:\begin{equation} \label{eqn:Frangi}
V_0(\sigma) = \left\{\begin{array}{ll}
0, & \text{ if } \lambda_2 < 0,\\ 
\exp(-\frac{R_B^{2}}{2\beta ^{2}})(1-\exp(-\frac{S^2}{2c^2})), & \text{ otherwise,}
\end{array}
\right.
\end{equation} where $|\lambda_1|<|\lambda_2|$ are the eigenvalues, $S = \sqrt{\lambda_1^2 + \lambda_2^2}$ is the second order structureness, $R_B = \frac{\|\lambda_1\|}{\|\lambda_2\|}$ is the blobness measure,  $\beta, c$ are image-dependent parameters for blobness and structureness terms, and $V_0$ stands for the vesselness value.

The entire multi-scale framework of Frangi filter can be mapped onto a neural network architecture \cite{ArXivWeilin}. In Frangi-Net, each step of the Frangi filter is replaced with a network counterpart and data normalization layers are added to enable end-to-end training. Multi-scale analysis is formed as a series of trainable filters, followed by eigenvalue computation in specialized fixed function network blocks. This is followed by another fixed function -- the actual vesselness measure as described in Eq.~\ref{eqn:Frangi}. 

\begin{figure*}[tbp]
  \centering
  \includegraphics[width=0.9\linewidth]{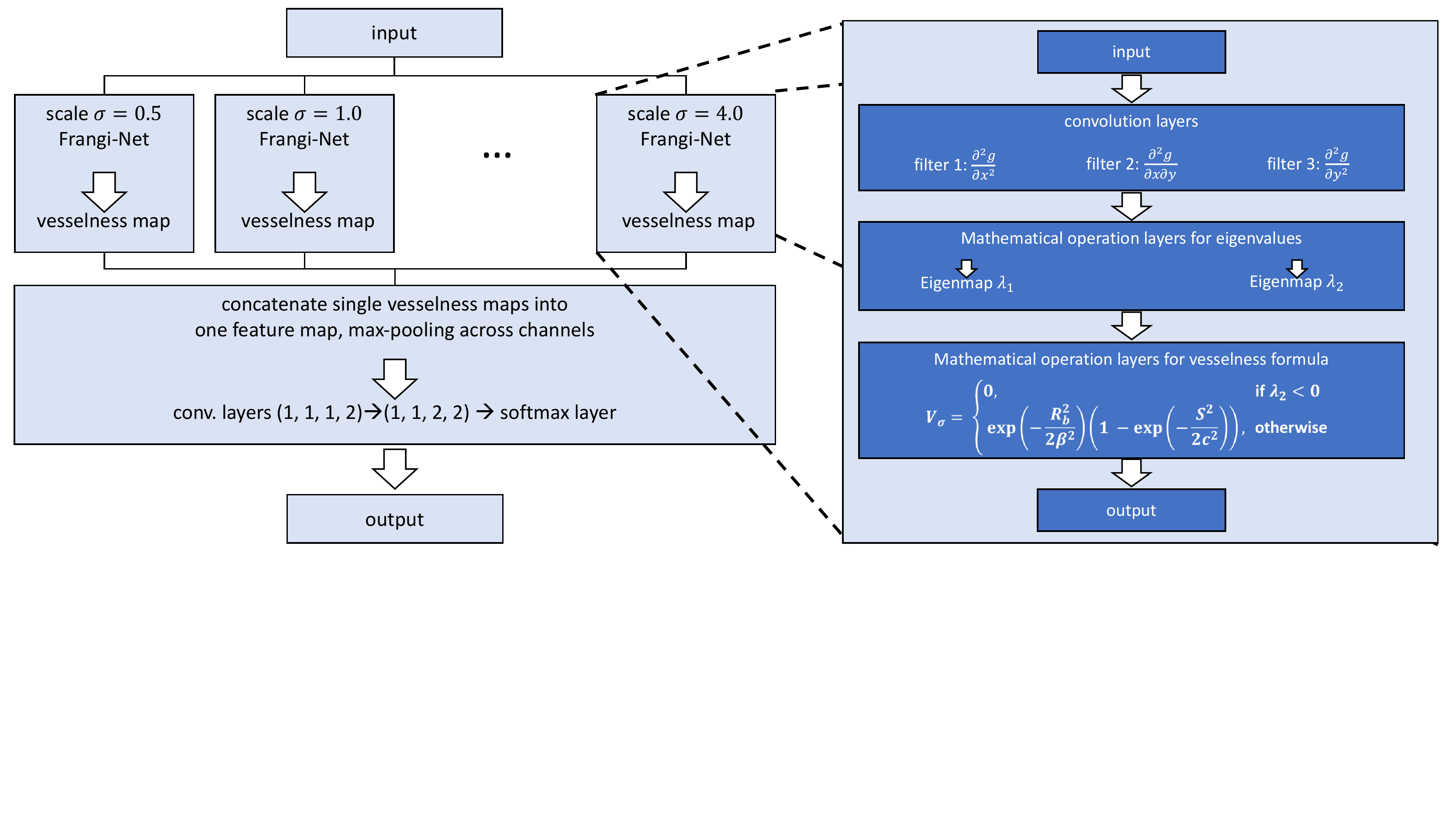}\vspace{-1.5cm}
 \caption{Architecture of Frangi-Net over 8 scales \change{$\sigma$}: \change{For each} single-scale \change{a} Frangi-Net \change{computes spatial derivatives $\frac{\partial^2 g}{\partial x^2}$, $\frac{\partial^2 g}{\partial x \partial y}$,  and $\frac{\partial^2 g}{\partial y^2}$. These are used to form a Hessian matrix of which eigenvalues $\lambda_1$ and $\lambda_2$ are extracted. Both are used to compute structureness $S$ and blobness $R_b$ which are required to compute the final vesselness at each pixel $V_\sigma$}. Image adapted from~\cite{ArXivWeilin}.}
   \label{fig:fmlpnet}
\end{figure*}


We compare the segmentation result of the proposed Frangi-Net with the original Frangi filter, and show that the Frangi-Net outperforms Frangi filter regarding all evaluation metrics. In comparison to the state-of-the-art image segmentation model U-Net, Frangi-Net contains less than 6\,\% the number of trainable parameters, while achieving an AUC score around 0.960, which is only 1\,\% inferior to that of the U-Net. \change{Adding a trainable guided-filter before Frangi-Net as preprocessing step yields an AUC 0.972 with only 8.5\,\% of the trainable parameters of U-Net which is statistically not distinguishable from U-Net's AUC of 0.974.}

 Hence using our approach of known operators, we are able to augment heuristic methods by blending them with methods of deep learning saving many trainable parameters.

\subsection{Deriving Networks}
\label{sec:rebinning}

\newcommand{\projcb}{{\bf A}_\text{CB}}
\newcommand{\projpb}{{\bf A}_\text{PB}}
\newcommand{\pcb}{{\bf p}_\text{CB}}
\newcommand{\ppb}{{\bf p}_\text{PB}}
\newcommand{\vol}{{\bf v}}

A third application of known operator learning that we would like to highlight in this paper, is the derivation of new network architectures from mathematical relations of the signal processing problem at hand. In the following, we are interested in hybrid imaging of magnetic resonance imaging (MRI) and X-ray imaging simultaneously. One major problem is that MRI $k$-space acquisitions typically allow parallel projection geometries, i.\,e. a line through the center $k$-space, while X-rays are associated with a divergent geometry such as fan- or cone-beam geometries. Both modalities allow different contrast mechanisms and simultaneous acquisition and overlay in the same image would be highly desirable for improved interventional guidance.

In the following, we assume to have sampled MRI projections $\x$ in $k$-space. By inverse Fourier Transform $\ifft$, they can be transformed into parallel projections $\ppb = \ifft \x$. Both parallel and cone-beam projections $\pcb$ are related to the volume under consideration $\vol$ by associated projection operations $\projpb$ and $\projcb$:
\begin{eqnarray}
\ppb = \projpb \vol \label{eq:ppb}\\
\pcb = \projcb \vol\label{eq:pcb}
\end{eqnarray}
As $\vol$ appears in both relations, we can solve Eq.~\ref{eq:ppb} for $\vol$ using the Moore-Penrose Pseudo Inverse:
$$\vol = \projpb^\top(\projpb\projpb^\top)^{-1} \ppb = \projpb^\top(\projpb\projpb^\top)^{-1} \ifft \x$$
Next, we can use $\vol$ in Eq.~\ref{eq:pcb} to yield
$$\pcb = \projcb  \projpb^\top(\projpb\projpb^\top)^{-1} \ifft \x. $$
Note that all operations on the path from $k$-space to $\pcb$ are known. Yet, $(\projpb\projpb^\top)^{-1}$ is expensive to determine and may need significant amounts of memory. As we know from reconstruction theory, this matrix often takes the form of a circulant matrix, i.\,e. a convolution. As such, we can approximate it with the chain of operations $\ifft\convfft\fft$ where $\convfft$ is a diagonal matrix. In order to add a few more degrees of freedom, we further add another diagonal operator in spatial domain $\weights$ to yield
\begin{equation}
\pcb = \projcb  \projpb^\top \weights \ifft\convfft\fft   \ifft \x = \projcb  \projpb^\top \weights \ifft\convfft \x
\end{equation}
as parallel to cone rebinning formula. In this formulation, only $\convfft$ and $\weights$ are unknown and need to be trained. By design both matrices are diagonal and therewith only have few unknown parameters.

Even though the training was conducted merely on numerical phantoms we can apply the learned algorithm on data acquired with a real MRI system without any loss of generality. Using only 15 parallel-beam MR projections we were able to compute a stacked fan-beam projection with both approaches. In Figure~\ref{fig:rebin_real_mri} the results of the analytical and learned algorithms are shown. The result of the learned algorithm has much sharper visual impression compared to the analytical approach which intrinsically suffers from ray-by-ray interpolation and thus from a blurring effect. \change{Note that additional smoothing could be incorporated into the network by regularization of the filter or additional hard-coded filter steps at request.}

\begin{figure}[tbp]
\begin{center}
  \centering
  \includegraphics[width=0.75\linewidth]{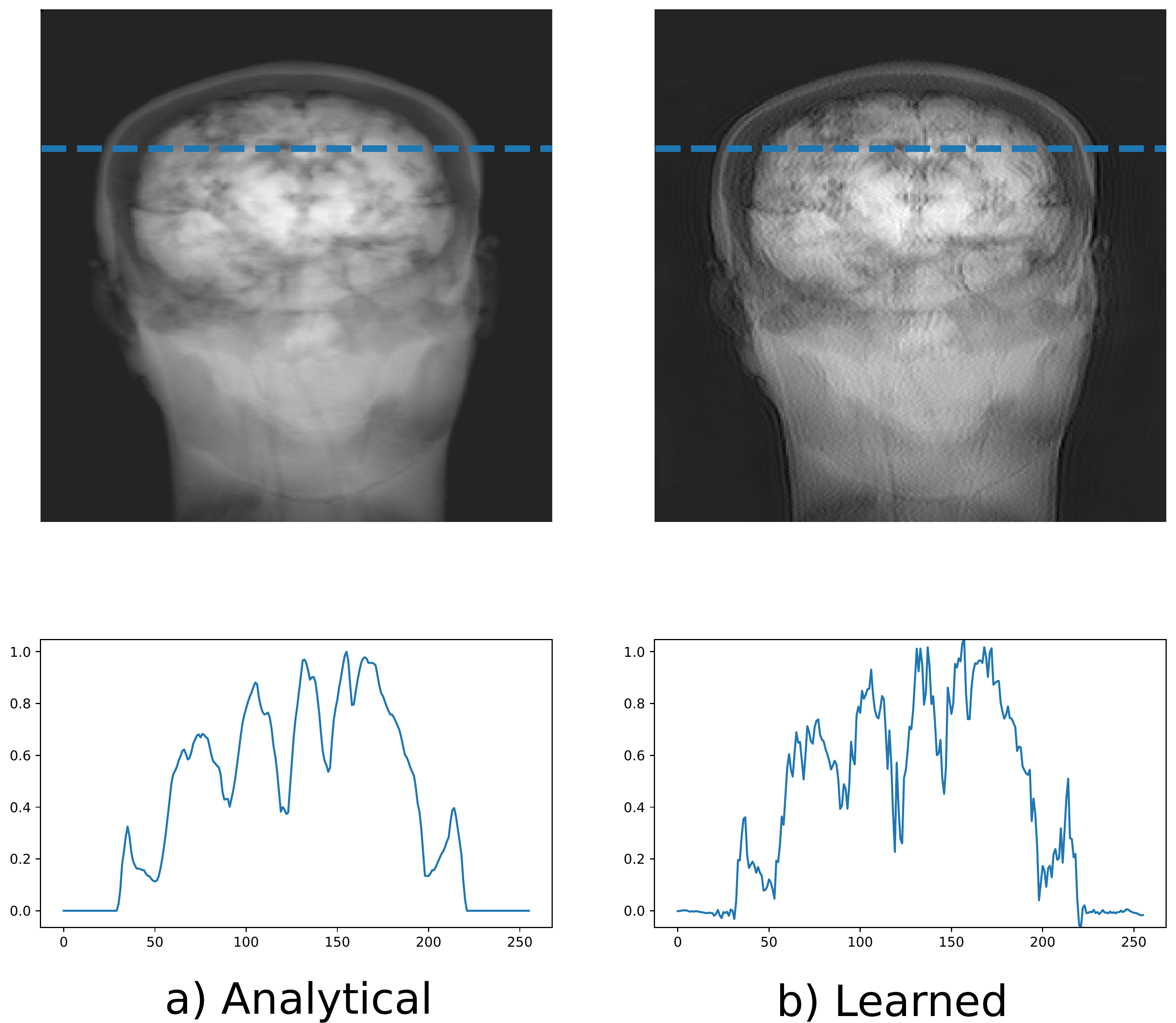}
\caption{\changefinal{Classical analytical rebinning vs.~derived neural networks:} The trained rebinning algorithm can directly applied to real MR projection data. Parallel-beam MR projection data is rebinned to a stacked fan-beam geometry with the analytical (a) and the learned algorithm (b). Note that the result of the learned method is much sharper as it avoids ray-by-ray interpolation.
}
\label{fig:rebin_real_mri}
\end{center}
\end{figure}

\section{Discussion}

For many applications, we do not know which operation is required in the ideal processing pipeline. Most machine learning tasks focus either on perceptual problems or man-made classes. Therefore, we only have limited knowledge on the ideal processing chain. In many cases, the human brain seems to have identified suitable solutions. Yet, our knowledge of the human brain is incomplete and search for well-suited deep architectures is a process of trial and error. Still, deep learning has shown to be able to solve tasks that were deemed as hard or close to impossible \cite{silver2016mastering}.

Now that deep learning also starts addressing fields of physics and classical signal processing, we are entering areas in which we have much better understanding of the underlying processes and therefore know that kind of mathematical operations need to be present in order to solve specific problems. Yet, during the derivation of our mathematical models, we often introduce simplifications that allow more compact descriptions and a more elegant solution. Still these simplifications introduce slight errors along the way and are often compensated using heuristic correction methods \cite{zhu2018image}.

In this paper, we have shown that inclusion of known operators is beneficial in terms of maximal error bounds. We demonstrated that in all cases in which we are able to use partial knowledge on the function at hand, the maximal errors that may remain after training of the network are reduced \changefinal{even} for networks of arbitrary depth. \changefinal{Note that in the future tighter error bounds than the ones described in this work might be identified that are independent of the use of known operators. Yet, our error analysis is still useful, as for the case of increasing number of known operations in the network, the magnitude of the bound shrinks up to the point of identity, if all operations are known.} To the knowledge of the authors, this is the first paper to attempt such a theoretical analysis of the use of known operators in neural network training.

In our experiments with CT reconstruction, we could demonstrate that we are able to tackle limited angle reconstructions using a standard filtered back-projection-type of algorithm. In fact, we only adopted weights while run-time, behaviour, and computational complexity remained unchanged. As we can map the trained algorithm back onto its original interpretation, we could also investigate shape and function of the learned weights. They demonstrated similarity to a heuristic method that could previously only be explained by intuition rather than by showing optimality. For the case, of our trained weights, we can demonstrate that they are optimal with respect to the training data.

Based on Frangi's vesselness, we could develop a trainable network for vessel detection. In our experiments, we could demonstrate that training of this net already yields improved filters for vessel detection that are close in terms of performance with a much more complex U-Net. \change{Further inclusion of a trainable denoising step yielded an accuracy that is statistically not distinguishable from U-Net.}

As last application of our approach, we investigated rebinning of MR data to a divergent beam geometry. For this kind of rebinning procedure, a fast convolution-based algorithm was previously unknown. Prior approaches relied on ray-by-ray interpolation that is typically introducing blurring. With our hypothesis that the inverse matrix operator takes the form of a circulant matrix in spatial domain in combination with an additional multiplicative weight, we could train a new algorithm successfully. The new approach is not just computationally efficient, it also features images of a degree of sharpness that was previously not reported in literature.

Although only applications from the medical domain are shown in this paper, this does not limit the generality of our theoretical analysis. Similar problems are found in many fields, e.\,g. computer vision \cite{univis91701757}, image super resolution \cite{univis91673802}, or audio signal processing \cite{univis91885067}.

Obviously, known operators have been embedded into neural networks already for a long time\change{. Already, LeCun et al. \cite{lecun1995convolutional} suggested convolution and pooling operators.} \change{Janderberg et al. introduced differentiable spatial transformations and their resampling into deep learning \cite{jaderberg2015spatial}. Lin et al. use this for image generation \cite{Lin_2018_CVPR}. Kulkarni et al. developed an entire deep convolutional graphics pipeline \cite{kulkarni2015deep}. Zhu et al. include differentiable projectors to disentangle 3D shape information from appearance \cite{zhu2018visual}. Tewari et al. integrate a differentiable model-based image generator to synthesize facial images \cite{Tewari_2017_ICCV}. Adler et al. shows an approach to partially learn the solution for ill-posed inverse problems\cite{adler2017solving}.} Ye et al. \cite{ye2018deep} introduced the Wavelet transform as multi-scale operator, Hammernik et al. \cite{hammernik2018learning} mapped entire energy minimization problems onto networks, and Wu et al. even included the guided filter as layer into a network \cite{DBLP:journals/corr/abs-1803-05619}. \change{As this list could be continued with many more references, we see} this as an emerging trend in deep learning. In fact, any operation that allows the computation of a sub-gradient \cite{rockafellar} is suited to be used in combination with the back-propagation algorithm. In order to integrate a new operator, only the partial derivatives / sub-gradients with respect to its inputs and its parameters have to be implemented. This allows inclusion of a wide range of operations. \change{To the best of our knowledge, this is the first paper giving a general argument for the effectiveness of such approaches.}

Next, the introduction of a known operator is also associated with a reduction of trainable parameters. We demonstrate this in this paper in all of our experiments. This allows us to work with much fewer training data and helps us to create models that can be transferred from synthetic training data to real measured data. Zarei et al. \cite{zarei2019user} drive this approach so far that they are able to train user-dependent image denoising filters using only few clicks from alternate forced-choice experiments. Thus, we believe that known operators may be a suitable approach to problems for which only limited data is available.

\change{At present we are unaware how to predict the benefit of using known operators before the actual experiment. Our analysis only focuses on maximum error bounds. Therefore, investigation of expected errors following for example the approach of Barron seems interesting for future work \cite{barron1994approximation}. Also analysis of the bias variance trade-off seems interesting. In \cite[Chapter 9]{duda2012pattern} Duda and Hart already hinted at the positive effect of prior knowledge on this trade-off.}

\begin{figure*}[tbp]
\begin{center}
\includegraphics[width=.9\linewidth]{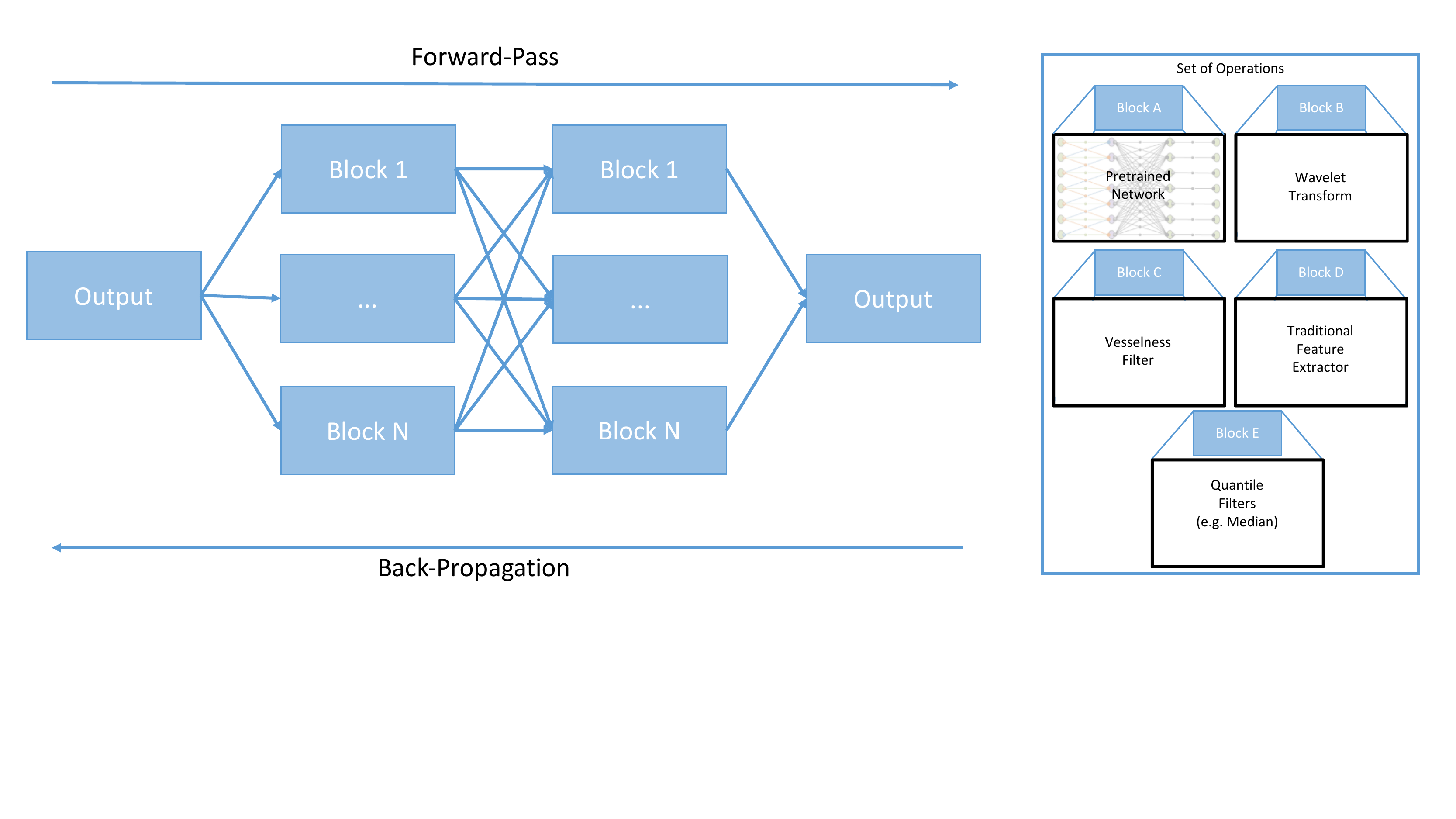}\vspace{-1.5cm}
\caption{\changefinal{Towards operator discovery and sequence analysis: } We hypothesise that Known Operator Learning may also be used to disentangle information efficiently. Offering several operators in parallel allows the network to find the best sequence of operations during the training process. In a subsequent step, blocks can be removed step-by-step to determine the minimal block networks.}
\label{fig:blocknet}
\end{center}
\end{figure*}

Lastly, we believe that known operators may be key in order to gain better understanding of deep networks. Similar to our experiments with Frangi-Net, we can start replacing layers with known operations and observe the effect on the performance of the network. \changefinal{From our theoretical analysis, we expect that inclusion of a known operation will not or only insignificantly reduce the system's performance.} This may allow us to find configurations for networks that only have few unknown operations while showing large parts that are explainable and understood. Figure~\ref{fig:blocknet} shows a variant of this process that is inspired by \cite{szegedy2015going}. Here, we offer a set of known operations in parallel and determine their optimal superposition by training of the network. In a second step, connections with low weights can be removed to iteratively determine the optimal sequence of operations. Furthermore, any known operator sequence can also be regarded as a hypothesis for a suitable algorithm for the problem at hand. By training, we are able to validate of falsify our hypothesis similar to our example of the derivation of a new network architecture.  

\section{Conclusion}

We believe that the use of known operators in deep networks is a \change{promising method}. In this paper, we demonstrate that the use of such reduces maximal error bounds and \change{experimentally show an reduction in} the number of trainable parameters. \change{Furthermore}, we \change{applied this to} the case of learning CT reconstruction yielding networks that are interpretable and that can be analysed with classical signal processing tools. Also mixing of deep and known operator learning is beneficial, as it allows us to build smaller networks with only 6\,\% of the parameters of a competing U-Net while being close with respect to their performance. Lastly, the known operators can also be found using mathematical derivation of networks. While keeping large parts of the mathematical operations, we only replace inefficient or unknown operations with deep learning techniques to find entirely new imaging formulas. While all of the applications shown in this paper stem only from the medical domain, we believe that this approach is applicable to all fields of physics and signal processing which is the focus of our future work. 

\section*{Acknowledgment}
The research leading to these results has received funding from the European Research Council (ERC) under the European Union's Horizon 2020 research and innovation program (ERC Grant No. 810316).

\change{
\section*{Author Contributions}
Andreas Maier is the main author of the paper and is responsible for the writing of the manuscript, theoretical analysis, and experimental design. Christopher Syben and Bernhard Stimpel contributed to the writing of Section~\ref{sec:rebinning} and the supporting experiments. Tobias W{\"urfl} and Mathis Hoffmann supported writing Section~\ref{sec:dlct} and performed the experiments reported in this section. Frank Schebesch contributed to the mathematical analysis and the writing thereof. Weilin Fu conducted the experiments supporting Section~\ref{sec:vessel} and contributed to their description. Leonid Mill, Lasse Kling, and Silke Christiansen contributed to the experimental design and the writing of the manuscript.
}

\change{
\section*{Data Availability Statement} 
All data in this publication are publicly available. The experiments in Section~\ref{sec:dlct} use data of the low-dose CT challenge \cite{mccollough2016tu}. Section~\ref{sec:vessel} uses the DRIVE database \cite{staal2004ridge}. The data for Section~\ref{sec:rebinning} is available in a Code Ocean Capsule available at \url{https://doi.org/10.24433/CO.8086142.v2}~\cite{weilin-codeocean}.
}

\change{
\section*{Code Availability Statement}
The code and data for this article, along with an accompanying computational environment, are available and executable online as a Code Ocean Capsule. Experiments in Section~\ref{sec:dlct} can be found at \url{https://doi.org/10.24433/CO.2164960.v1}~\cite{mathis-codeocean}. The code on learning vesselness in Section~\ref{sec:vessel} are published at \url{https://doi.org/10.24433/CO.5016803.v2}~\cite{weilin-codeocean}. Code for Section~\ref{sec:rebinning} is available at \url{https://doi.org/10.24433/CO.8086142.v2}~\cite{syben-codeocean}. 
The code capsules for the experiments in Section~\ref{sec:dlct} and Section~\ref{sec:rebinning}  were implemented using the open source framework PYRO-NN \cite{syben2019pyro}.
}



%
\bibliographystyle{naturemag}

\begin{thebibliography}{10}
\expandafter\ifx\csname url\endcsname\relax
  \def\url#1{\texttt{#1}}\fi
\expandafter\ifx\csname urlprefix\endcsname\relax\def\urlprefix{URL }\fi
\providecommand{\bibinfo}[2]{#2}
\providecommand{\eprint}[2][]{\url{#2}}

\bibitem{niemann2013pattern}
\bibinfo{author}{Niemann, H.}
\newblock \emph{\bibinfo{title}{Pattern Analysis and Understanding}},
  vol.~\bibinfo{volume}{4} (\bibinfo{publisher}{Springer Science \& Business
  Media}, \bibinfo{year}{2013}).

\bibitem{lecun1995convolutional}
\bibinfo{author}{LeCun, Y.} \& \bibinfo{author}{Bengio, Y.}
\newblock \bibinfo{title}{Convolutional networks for images, speech, and time
  series}.
\newblock \emph{\bibinfo{journal}{The handbook of brain theory and neural
  networks}} \textbf{\bibinfo{volume}{3361}}, \bibinfo{pages}{1995}
  (\bibinfo{year}{1995}).

\bibitem{krizhevsky2012imagenet}
\bibinfo{author}{Krizhevsky, A.}, \bibinfo{author}{Sutskever, I.} \&
  \bibinfo{author}{Hinton, G.~E.}
\newblock \bibinfo{title}{Imagenet classification with deep convolutional
  neural networks}.
\newblock In \emph{\bibinfo{booktitle}{Advances in Neural Information
  Processing Systems}}, \bibinfo{pages}{1097--1105} (\bibinfo{year}{2012}).

\bibitem{lecun2015deep}
\bibinfo{author}{LeCun, Y.}, \bibinfo{author}{Bengio, Y.} \&
  \bibinfo{author}{Hinton, G.}
\newblock \bibinfo{title}{Deep learning}.
\newblock \emph{\bibinfo{journal}{Nature}} \textbf{\bibinfo{volume}{521}},
  \bibinfo{pages}{436} (\bibinfo{year}{2015}).

\bibitem{dong2014learning}
\bibinfo{author}{Dong, C.}, \bibinfo{author}{Loy, C.~C.}, \bibinfo{author}{He,
  K.} \& \bibinfo{author}{Tang, X.}
\newblock \bibinfo{title}{Learning a deep convolutional network for image
  super-resolution}.
\newblock In \emph{\bibinfo{booktitle}{European Conference on Computer
  Vision}}, \bibinfo{pages}{184--199} (\bibinfo{organization}{Springer},
  \bibinfo{year}{2014}).

\bibitem{xie2012image}
\bibinfo{author}{Xie, J.}, \bibinfo{author}{Xu, L.} \& \bibinfo{author}{Chen,
  E.}
\newblock \bibinfo{title}{Image denoising and inpainting with deep neural
  networks}.
\newblock In \emph{\bibinfo{booktitle}{Advances in Neural Information
  Processing Systems}}, \bibinfo{pages}{341--349} (\bibinfo{year}{2012}).

\bibitem{wang2018image}
\bibinfo{author}{Wang, G.}, \bibinfo{author}{Ye, J.~C.},
  \bibinfo{author}{Mueller, K.} \& \bibinfo{author}{Fessler, J.~A.}
\newblock \bibinfo{title}{Image reconstruction is a new frontier of machine
  learning.}
\newblock \emph{\bibinfo{journal}{IEEE Transactions on Medical Imaging}}
  \textbf{\bibinfo{volume}{37}}, \bibinfo{pages}{1289--1296}
  (\bibinfo{year}{2018}).

\bibitem{Cohen2018distribution}
\bibinfo{author}{Cohen, J.~P.}, \bibinfo{author}{Luck, M.} \&
  \bibinfo{author}{Honari, S.}
\newblock \bibinfo{title}{Distribution matching losses can hallucinate features
  in medical image translation}.
\newblock In \bibinfo{editor}{Frangi, A.~F.}, \bibinfo{editor}{Schnabel,
  J.~A.}, \bibinfo{editor}{Davatzikos, C.},
  \bibinfo{editor}{Alberola-L{\'o}pez, C.} \& \bibinfo{editor}{Fichtinger, G.}
  (eds.) \emph{\bibinfo{booktitle}{Medical Image Computing and Computer
  Assisted Intervention -- MICCAI 2018}}, \bibinfo{pages}{529--536}
  (\bibinfo{publisher}{Springer International Publishing},
  \bibinfo{address}{Cham}, \bibinfo{year}{2018}).

\bibitem{huang2018considerations}
\bibinfo{author}{Huang, Y.} \emph{et~al.}
\newblock \bibinfo{title}{Some investigations on robustness of deep learning in
  limited angle tomography}.
\newblock In \bibinfo{editor}{Frangi, A.~F.}, \bibinfo{editor}{Schnabel,
  J.~A.}, \bibinfo{editor}{Davatzikos, C.},
  \bibinfo{editor}{Alberola-L{\'o}pez, C.} \& \bibinfo{editor}{Fichtinger, G.}
  (eds.) \emph{\bibinfo{booktitle}{Medical Image Computing and Computer
  Assisted Intervention -- MICCAI 2018}}, \bibinfo{pages}{145--153}
  (\bibinfo{publisher}{Springer International Publishing},
  \bibinfo{address}{Cham}, \bibinfo{year}{2018}).

\bibitem{deeplearningct}
\bibinfo{author}{W{\"u}rfl, T.}, \bibinfo{author}{Ghesu, F.~C.},
  \bibinfo{author}{Christlein, V.} \& \bibinfo{author}{Maier, A.}
\newblock \bibinfo{title}{Deep learning computed tomography}.
\newblock In \emph{\bibinfo{booktitle}{International Conference on Medical
  Image Computing and Computer-Assisted Intervention}},
  \bibinfo{pages}{432--440} (\bibinfo{organization}{Springer},
  \bibinfo{year}{2016}).

\bibitem{ArXivWeilin}
\bibinfo{author}{Fu, W.} \emph{et~al.}
\newblock \bibinfo{title}{{Frangi-Net: A Neural Network Approach to Vessel
  Segmentation}}.
\newblock In \bibinfo{editor}{Maier, A.} \emph{et~al.} (eds.)
  \emph{\bibinfo{booktitle}{{Bildverarbeitung f{\"u}r die Medizin 2018}}},
  \bibinfo{pages}{341--346} (\bibinfo{year}{2018}).

\bibitem{maier2018precision}
\bibinfo{author}{Maier, A.} \emph{et~al.}
\newblock \bibinfo{title}{{Precision Learning: Towards Use of Known Operators
  in Neural Networks}}.
\newblock In \bibinfo{editor}{Tan, J. K.~T.} (ed.)
  \emph{\bibinfo{booktitle}{{2018 24rd International Conference on Pattern
  Recognition (ICPR)}}}, \bibinfo{pages}{183--188} (\bibinfo{year}{2018}).
\newblock
  \urlprefix\url{https://www5.informatik.uni-erlangen.de/Forschung/Publikationen/2018/Maier18-PLT.pdf}.

\bibitem{syben2018deriving}
\bibinfo{author}{Syben, C.} \emph{et~al.}
\newblock \bibinfo{title}{Deriving neural network architectures using precision
  learning: Parallel-to-fan beam conversion}.
\newblock In \emph{\bibinfo{booktitle}{German Conference on Pattern Recognition
  (GCPR)}} (\bibinfo{year}{2018}).

\bibitem{cybenko1989approximation}
\bibinfo{author}{Cybenko, G.}
\newblock \bibinfo{title}{Approximation by superpositions of a sigmoidal
  function}.
\newblock \emph{\bibinfo{journal}{Mathematics of Control, Signals and Systems}}
  \textbf{\bibinfo{volume}{2}}, \bibinfo{pages}{303--314}
  (\bibinfo{year}{1989}).

\bibitem{hornik1991approximation}
\bibinfo{author}{Hornik, K.}
\newblock \bibinfo{title}{Approximation capabilities of multilayer feedforward
  networks}.
\newblock \emph{\bibinfo{journal}{Neural networks}}
  \textbf{\bibinfo{volume}{4}}, \bibinfo{pages}{251--257}
  (\bibinfo{year}{1991}).

\bibitem{maier2019gentle}
\bibinfo{author}{Maier, A.}, \bibinfo{author}{Syben, C.},
  \bibinfo{author}{Lasser, T.} \& \bibinfo{author}{Riess, C.}
\newblock \bibinfo{title}{A gentle introduction to deep learning in medical
  image processing}.
\newblock \emph{\bibinfo{journal}{Zeitschrift f{\"u}r Medizinische Physik}}
  \textbf{\bibinfo{volume}{29}}, \bibinfo{pages}{86--101}
  (\bibinfo{year}{2019}).

\bibitem{parker1982optimal}
\bibinfo{author}{Parker, D.~L.}
\newblock \bibinfo{title}{Optimal short scan convolution reconstruction for fan
  beam ct}.
\newblock \emph{\bibinfo{journal}{Medical Physics}}
  \textbf{\bibinfo{volume}{9}}, \bibinfo{pages}{254--257}
  (\bibinfo{year}{1982}).

\bibitem{schafer2017modified}
\bibinfo{author}{Sch{\"a}fer, D.}, \bibinfo{author}{van~de Haar, P.} \&
  \bibinfo{author}{Grass, M.}
\newblock \bibinfo{title}{Modified parker weights for super short scan cone
  beam ct}.
\newblock In \emph{\bibinfo{booktitle}{Proc. 14th Int. Meeting Fully
  Three-Dimensional Image Reconstruction Radiol. Nucl. Med.}},
  \bibinfo{pages}{49--52} (\bibinfo{year}{2017}).

\bibitem{vesselness-frangi}
\bibinfo{author}{Frangi, A.~F.}, \bibinfo{author}{Niessen, W.~J.},
  \bibinfo{author}{Vincken, K.~L.} \& \bibinfo{author}{Viergever, M.~A.}
\newblock \bibinfo{title}{Multiscale vessel enhancement filtering}.
\newblock In \emph{\bibinfo{booktitle}{International Conference on Medical
  Image Computing and Computer-Assisted Intervention}},
  \bibinfo{pages}{130--137} (\bibinfo{organization}{Springer},
  \bibinfo{year}{1998}).

\bibitem{silver2016mastering}
\bibinfo{author}{Silver, D.} \emph{et~al.}
\newblock \bibinfo{title}{Mastering the game of go with deep neural networks
  and tree search}.
\newblock \emph{\bibinfo{journal}{Nature}} \textbf{\bibinfo{volume}{529}},
  \bibinfo{pages}{484} (\bibinfo{year}{2016}).

\bibitem{zhu2018image}
\bibinfo{author}{Zhu, B.}, \bibinfo{author}{Liu, J.~Z.},
  \bibinfo{author}{Cauley, S.~F.}, \bibinfo{author}{Rosen, B.~R.} \&
  \bibinfo{author}{Rosen, M.~S.}
\newblock \bibinfo{title}{Image reconstruction by domain-transform manifold
  learning}.
\newblock \emph{\bibinfo{journal}{Nature}} \textbf{\bibinfo{volume}{555}},
  \bibinfo{pages}{487} (\bibinfo{year}{2018}).

\bibitem{univis91701757}
\bibinfo{author}{F{\"u}rsattel, P.}, \bibinfo{author}{Plank, C.},
  \bibinfo{author}{Maier, A.} \& \bibinfo{author}{Riess, C.}
\newblock \bibinfo{title}{{Accurate Laser Scanner to Camera Calibration with
  Application to Range Sensor Evaluation}}.
\newblock \emph{\bibinfo{journal}{IPSJ Transactions on Computer Vision and
  Applications}} \textbf{\bibinfo{volume}{9}} (\bibinfo{year}{2017}).
\newblock
  \urlprefix\url{https://link.springer.com/article/10.1186/s41074-017-0032-5}.

\bibitem{univis91673802}
\bibinfo{author}{K{\"o}hler, T.} \emph{et~al.}
\newblock \bibinfo{title}{{Robust Multiframe Super-Resolution Employing
  Iteratively Re-Weighted Minimization}}.
\newblock \emph{\bibinfo{journal}{IEEE Transactions on Computational Imaging}}
  \textbf{\bibinfo{volume}{2}}, \bibinfo{pages}{42--58} (\bibinfo{year}{2016}).
\newblock
  \urlprefix\url{https://www5.informatik.uni-erlangen.de/Forschung/Publikationen/2016/Kohler16-RMS.pdf}.

\bibitem{univis91885067}
\bibinfo{author}{Aubreville, M.} \emph{et~al.}
\newblock \bibinfo{title}{{Deep Denoising for Hearing Aid Applications}}.
\newblock In \bibinfo{editor}{IEEE} (ed.) \emph{\bibinfo{booktitle}{{16th
  International Workshop on Acoustic Signal Enhancement (IWAENC)}}},
  \bibinfo{pages}{361--365} (\bibinfo{year}{2018}).

\bibitem{jaderberg2015spatial}
\bibinfo{author}{Jaderberg, M.}, \bibinfo{author}{Simonyan, K.},
  \bibinfo{author}{Zisserman, A.} \& \bibinfo{author}{Kavukcuoglu, K.}
\newblock \bibinfo{title}{Spatial transformer networks}.
\newblock In \emph{\bibinfo{booktitle}{Advances in Neural Information
  Processing Systems}}, \bibinfo{pages}{2017--2025} (\bibinfo{year}{2015}).

\bibitem{Lin_2018_CVPR}
\bibinfo{author}{Lin, C.-H.}, \bibinfo{author}{Yumer, E.},
  \bibinfo{author}{Wang, O.}, \bibinfo{author}{Shechtman, E.} \&
  \bibinfo{author}{Lucey, S.}
\newblock \bibinfo{title}{St-gan: Spatial transformer generative adversarial
  networks for image compositing}.
\newblock In \emph{\bibinfo{booktitle}{The IEEE Conference on Computer Vision
  and Pattern Recognition (CVPR)}} (\bibinfo{year}{2018}).

\bibitem{kulkarni2015deep}
\bibinfo{author}{Kulkarni, T.~D.}, \bibinfo{author}{Whitney, W.~F.},
  \bibinfo{author}{Kohli, P.} \& \bibinfo{author}{Tenenbaum, J.}
\newblock \bibinfo{title}{Deep convolutional inverse graphics network}.
\newblock In \emph{\bibinfo{booktitle}{Advances in Neural Information
  Processing Systems}}, \bibinfo{pages}{2539--2547} (\bibinfo{year}{2015}).

\bibitem{zhu2018visual}
\bibinfo{author}{Zhu, J.-Y.} \emph{et~al.}
\newblock \bibinfo{title}{Visual object networks: Image generation with
  disentangled 3d representations}.
\newblock In \emph{\bibinfo{booktitle}{Advances in Neural Information
  Processing Systems}}, \bibinfo{pages}{118--129} (\bibinfo{year}{2018}).

\bibitem{Tewari_2017_ICCV}
\bibinfo{author}{Tewari, A.} \emph{et~al.}
\newblock \bibinfo{title}{Mofa: Model-based deep convolutional face autoencoder
  for unsupervised monocular reconstruction}.
\newblock In \emph{\bibinfo{booktitle}{The IEEE International Conference on
  Computer Vision (ICCV) Workshops}} (\bibinfo{year}{2017}).

\bibitem{adler2017solving}
\bibinfo{author}{Adler, J.} \& \bibinfo{author}{{\"O}ktem, O.}
\newblock \bibinfo{title}{Solving ill-posed inverse problems using iterative
  deep neural networks}.
\newblock \emph{\bibinfo{journal}{Inverse Problems}}
  \textbf{\bibinfo{volume}{33}}, \bibinfo{pages}{124007}
  (\bibinfo{year}{2017}).

\bibitem{ye2018deep}
\bibinfo{author}{Ye, J.~C.}, \bibinfo{author}{Han, Y.} \& \bibinfo{author}{Cha,
  E.}
\newblock \bibinfo{title}{Deep convolutional framelets: A general deep learning
  framework for inverse problems}.
\newblock \emph{\bibinfo{journal}{SIAM Journal on Imaging Sciences}}
  \textbf{\bibinfo{volume}{11}}, \bibinfo{pages}{991--1048}
  (\bibinfo{year}{2018}).

\bibitem{hammernik2018learning}
\bibinfo{author}{Hammernik, K.} \emph{et~al.}
\newblock \bibinfo{title}{Learning a variational network for reconstruction of
  accelerated mri data}.
\newblock \emph{\bibinfo{journal}{Magnetic Resonance in Medicine}}
  \textbf{\bibinfo{volume}{79}}, \bibinfo{pages}{3055--3071}
  (\bibinfo{year}{2018}).

\bibitem{DBLP:journals/corr/abs-1803-05619}
\bibinfo{author}{Wu, H.}, \bibinfo{author}{Zheng, S.}, \bibinfo{author}{Zhang,
  J.} \& \bibinfo{author}{Huang, K.}
\newblock \bibinfo{title}{Fast end-to-end trainable guided filter}.
\newblock \emph{\bibinfo{journal}{CoRR}}
  \textbf{\bibinfo{volume}{abs/1803.05619}} (\bibinfo{year}{2018}).
\newblock \urlprefix\url{http://arxiv.org/abs/1803.05619}.
\newblock \eprint{1803.05619}.

\bibitem{rockafellar}
\bibinfo{author}{Rockafellar, R.}
\newblock \emph{\bibinfo{title}{Convex Analysis}}.
\newblock Princeton landmarks in mathematics and physics
  (\bibinfo{publisher}{Princeton University Press}, \bibinfo{year}{1970}).
\newblock \urlprefix\url{https://books.google.de/books?id=1TiOka9bx3sC}.

\bibitem{zarei2019user}
\bibinfo{author}{Zarei, S.}, \bibinfo{author}{Stimpel, B.},
  \bibinfo{author}{Syben, C.} \& \bibinfo{author}{Maier, A.}
\newblock \bibinfo{title}{{User Loss A Forced-Choice-Inspired Approach to Train
  Neural Networks Directly by User Interaction}}.
\newblock In \emph{\bibinfo{booktitle}{{Bildverarbeitung f{\"u}r die Medizin
  2019}}}, {Informatik aktuell}, \bibinfo{pages}{92--97}
  (\bibinfo{year}{2019}).
\newblock
  \urlprefix\url{https://www5.informatik.uni-erlangen.de/Forschung/Publikationen/2019/Zarei19-ULA.pdf}.

\bibitem{barron1994approximation}
\bibinfo{author}{Barron, A.~R.}
\newblock \bibinfo{title}{Approximation and estimation bounds for artificial
  neural networks}.
\newblock \emph{\bibinfo{journal}{Machine learning}}
  \textbf{\bibinfo{volume}{14}}, \bibinfo{pages}{115--133}
  (\bibinfo{year}{1994}).

\bibitem{duda2012pattern}
\bibinfo{author}{Duda, R.~O.}, \bibinfo{author}{Hart, P.~E.} \&
  \bibinfo{author}{Stork, D.~G.}
\newblock \emph{\bibinfo{title}{Pattern classification}}
  (\bibinfo{publisher}{John Wiley \& Sons}, \bibinfo{year}{2012}).

\bibitem{szegedy2015going}
\bibinfo{author}{Szegedy, C.} \emph{et~al.}
\newblock \bibinfo{title}{Going deeper with convolutions}.
\newblock In \emph{\bibinfo{booktitle}{2015 IEEE Conference on Computer Vision
  and Pattern Recognition (CVPR)}}, \bibinfo{pages}{1--9}
  (\bibinfo{year}{2015}).

\bibitem{mccollough2016tu}
\bibinfo{author}{McCollough, C.}
\newblock \bibinfo{title}{Tu-fg-207a-04: Overview of the low dose ct grand
  challenge}.
\newblock \emph{\bibinfo{journal}{Medical Physics}}
  \textbf{\bibinfo{volume}{43}}, \bibinfo{pages}{3759--3760}
  (\bibinfo{year}{2016}).

\bibitem{staal2004ridge}
\bibinfo{author}{Staal, J.}, \bibinfo{author}{Abr{\`a}moff, M.~D.},
  \bibinfo{author}{Niemeijer, M.}, \bibinfo{author}{Viergever, M.~A.} \&
  \bibinfo{author}{Van~Ginneken, B.}
\newblock \bibinfo{title}{Ridge-based vessel segmentation in color images of
  the retina}.
\newblock \emph{\bibinfo{journal}{{IEEE Transactions on Medical Imaging}}}
  \textbf{\bibinfo{volume}{23}}, \bibinfo{pages}{501--509}
  (\bibinfo{year}{2004}).

\bibitem{weilin-codeocean}
\bibinfo{author}{Fu, W.}
\newblock \bibinfo{title}{Frangi-net on high-resolution fundus ({HRF}) image
  database}.
\newblock \emph{\bibinfo{journal}{Code Ocean}}  (\bibinfo{year}{2019}).
\newblock \bibinfo{note}{\url{https://doi.org/10.24433/CO.5016803.v2}}.

\bibitem{mathis-codeocean}
\bibinfo{author}{Syben, C.} \& \bibinfo{author}{Hoffmann, M.}
\newblock \bibinfo{title}{Learning {CT} reconstruction}.
\newblock \emph{\bibinfo{journal}{Code Ocean}}  (\bibinfo{year}{2019}).
\newblock \bibinfo{note}{\url{https://doi.org/10.24433/CO.2164960.v1}}.

\bibitem{syben-codeocean}
\bibinfo{author}{Syben, C.}
\newblock \bibinfo{title}{Deriving neural networks}.
\newblock \emph{\bibinfo{journal}{Code Ocean}}  (\bibinfo{year}{2019}).
\newblock \bibinfo{note}{\url{https://doi.org/10.24433/CO.8086142.v2}}.

\bibitem{syben2019pyro}
\bibinfo{author}{Syben, C.} \emph{et~al.}
\newblock \bibinfo{title}{{PYRO-NN}: Python reconstruction operators in neural
  networks}.
\newblock \emph{\bibinfo{journal}{arXiv preprint arXiv:1904.13342}}
  (\bibinfo{year}{2019}).

\end{thebibliography}

%






\end{document}